\pdfoutput=1

\documentclass[final,5p,times,twocolumn]{elsarticle}

\usepackage{amssymb}
\usepackage{amsmath}

\usepackage{lineno}

\usepackage[hidelinks]{hyperref}
\usepackage{url}
\usepackage{makecell}
\usepackage{booktabs}
\usepackage{multirow}
\usepackage{array}
\usepackage{threeparttable}
\usepackage{graphicx}
\usepackage{subcaption}
\usepackage{pdflscape}
\usepackage{tabularx}
\usepackage{array}
\usepackage{setspace}
\usepackage{adjustbox}

\makeatletter
\def\ps@pprintTitle{%
  \let\@oddhead\@empty
  \let\@evenhead\@empty
  \def\@oddfoot{\centerline{\thepage}}%
  \let\@evenfoot\@oddfoot}
\makeatother

\begin{document}
\hyphenation{SIFTING}

\begin{frontmatter}

\title{SIFTING: A Novel LLM-Based Framework for Structured and Transparent Information Extraction from Clinical Free-Text Reports, with Application to Tumor Staging in Lung Cancer}

\author[axana]{Hess M.\corref{cor1}}
\ead{mirco@axana.com}

\author[axana]{van Veenendaal G.} 
\author[axana]{Wakkie, J.}
\author[uclh, ucl]{Yiwen, S.}
\author[mseft]{Lawson, M. H.} 
\author[glasgow]{Maclay, J. D.} 
\author[uclh, ucl]{Nair, A.} 
\author[uclh,ucl]{Navani, N.} 

\cortext[cor1]{Corresponding author.}

\affiliation[axana]{organization={Axana B.V.},
            addressline={Plantage Kerklaan 18-0}, 
            city={Amsterdam},
            postcode={1018~TB}, 
            state={North Holland},
            country={Netherlands}}

\affiliation[uclh]{organization={University College London Hospitals NHS Foundation Trust}, city={London}, country={United Kingdom}}
\affiliation[ucl]{organization={University College London }, city={London}, country={United Kingdom}}
\affiliation[mseft]{organization={Mid \& South Essex NHS Foundation Trust}, city={Essex}, country={United Kingdom}}
\affiliation[glasgow]{organization={NHS Greater Glasgow and Clyde}, city={Glasgow}, country={United Kingdom}}


\begin{abstract}
\paragraph{\textbf{Background}}
Large language models (LLMs) show promise for extracting information from clinical free-text documents, but their outputs are often unstructured and lack traceability, complicating validation and adoption in clinical workflows.
In this work we introduce SIFTING, an LLM-based framework designed to address these shortcomings.

\paragraph{\textbf{Methods}}
SIFTING combines the language comprehension capabilities of LLMs with segment-level processing and structured prompts with strict output control, linking findings to the source text to enable both accurate and transparent information extraction.
To demonstrate its capabilities, we applied the framework to the task of extracting tumor T-stage information from 130 lung cancer radiology reports (SIFTING-T-stage). 
A compact 4-bit quantized version of the open-source LLM Llama-3.3-70B (35~GB) was used in a fully self-hosted setup, providing full control over data and model.
Performance was evaluated against a reference standard created by four clinical experts and compared with a range of LLMs as used in a conventional single-prompt approach, using bootstrap resampling to estimate confidence intervals.

\paragraph{\textbf{Results}}
SIFTING-T-stage achieved an accuracy of 90\% (95\%~CI:~84–95) against the reference standard.
We found its performance to be comparable to even the largest state-of-the-art LLMs with reasoning capabilities and to be interchangeable with clinical experts (p $<$ 0.001), while at the same time offering full traceability through source text references.

\paragraph{\textbf{Conclusion}}
SIFTING enables accurate, structured, and traceable information extraction from clinical free-text documents. 
It ensures data control, reproducibility, and verifiable outputs that can support clinical validation and workflow integration.
\newline

\end{abstract}



\end{frontmatter}
\thispagestyle{empty}



\section{Introduction}
\subsection{Background}

In healthcare, a substantial proportion of clinical information is recorded in free-text documents rather than structured fields, as such narratives can provide a more comprehensive picture than coded data alone.
Examples include patient letters, radiology reports, pathology findings, and discharge summaries, which can contain information essential for patient care, research, and healthcare management \cite{ford_should_2020, seinen_using_2025}. 
However, the unstructured nature of these texts makes it challenging to extract specific details at scale, particularly when reports are lengthy, inconsistent, or numerous enough to create information overload in a time-constrained setting \cite{wallis_radiology_2011, lukaszewicz_art_2016}. 
Inefficient or inaccurate extraction not only increases the burden on clinical workflows but may also contribute to error, highlighting the need for more reliable and scalable information extraction methods \cite{piotrkowicz_finding_2019}.
For such methods to be trusted and adopted in clinical applications, they must combine accuracy with transparency and provide mechanisms for clinician oversight and manual verification \cite{akinci_dantonoli_large_2024, nasarian_designing_2024}.

Natural language processing (NLP) and, more recently, large language models (LLMs) have emerged as powerful tools to automatically extract and interpret information from text.
While traditional NLP techniques have shown feasibility, they are often rule-based and lack the robustness needed to handle complex and variable medical language and document structures \cite{huesch_evaluating_2018, landolsi_information_2023}.
Recent studies demonstrate the potential of LLMs to improve upon such information extraction systems, as they offer a more flexible and generalizable alternative to traditional NLP methods \cite{peng_study_2025, hsu_deep_2022, joseph-thomas_comparing_2025}.
Their extensive pre-training on vast datasets enables them to better accommodate variations in reporting style and interpret complex linguistic nuances, making them well-suited candidates for clinical applications \cite{maity_large_2025, kim_large_2024, can_large_2024, chen_large_2025}. 
However, LLMs also present challenges, including the risk of hallucination, where the model produces information that appears credible but is inaccurate or unsubstantiated, as well as limited interpretability and a lack of traceability to the original source text \cite{casey_systematic_2021, mcnicholas_natural_2025}.

\subsection{Contribution}
In this work we introduce SIFTING (A Scalable and Intelligent Framework for Transparent Information Gathering), a novel framework that leverages the strengths of LLMs for gathering information from unstructured text with high accuracy while also providing full traceability and reducing the risk of hallucinations by design.
To demonstrate the capabilities of SIFTING, we applied the framework to the task of extracting tumor information from radiology reports to determine the tumor stage.
These reports are a primary means of communication between radiologists and referring clinicians, conveying essential information for patient management and treatment decisions.
The 8th edition of the TNM Classification of Malignant Tumors provides a standardized guideline for lung cancer staging, categorizing tumors based on size, local invasion, lymph node involvement, and metastases \cite{brierley_j_tnm_2017}.
Applying this guideline requires extracting precise quantitative features (e.g. tumor size in millimeters) as well as clinically relevant qualitative findings (e.g. invasion into adjacent structures) from the reports.
Reports often vary in structure and terminology and may contain ambiguous or interpretive language \cite{krupinski_influence_2012}.
This combination of quantitative and qualitative information requirements, together with the need for verifiability in complex cases, makes T-staging based on radiology reports an ideal task for evaluating and showcasing the SIFTING framework.

\section{Methods}
\subsection{Dataset}

A total of 130 chest-abdomen-pelvis (CAP) radiology reports were obtained from a commercial data provider (Segmed, Inc.) for inclusion in this study as an external validation set.
All reports corresponded to patients with a confirmed diagnosis of primary non-small cell lung cancer (NSCLC) who underwent imaging specifically for staging.
Each report followed a structured format with sections such as clinical details, report, modality described, body part, and impression.
This organization aligns with the American College of Radiology (ACR) practice guideline, which recommends placing descriptive findings in the body of the report and reserving diagnostic statements for the impression section \cite{gershanik_critical_2011}.
The reports were fully de-identified, permitting their use for research purposes without additional patient consent or institutional review board approval.
Patient characteristics are provided in Table \ref{table_characteristics}.

\newcolumntype{L}[1]{>{\raggedright\arraybackslash}p{#1}} 
\newcolumntype{Y}{>{\raggedright\arraybackslash}X}        

\begin{table}[htbp]
\centering
\begin{threeparttable}
\caption{Patient Characteristics}
\label{table_characteristics}
\vspace{1mm}
\begin{tabularx}{\columnwidth}{L{2cm} L{4.5cm} Y }
\toprule
Characteristic & Value & Count (\%) \\
\midrule
\multirow[t]{4}{*}{\makecell[tl]{Patient age \\ (years)}}
 & $<$~60     & 21 (16) \\
 & 60--69     & 46 (35) \\
 & 70--79     & 46 (35) \\
 & $\geq$~80 & 17 (13) \\
\addlinespace
\multirow[t]{2}{*}{Sex} 
 & Female & 74 (57) \\
 & Male   & 56 (43) \\
\addlinespace
\multirow[t]{3}{*}{Race} 
 & White                   & 101 (78) \\
 & Black / African American & 25 (19) \\
 & Asian                   & 4 (3) \\
\bottomrule
\end{tabularx}
\begin{tablenotes}[flushleft]
\item \footnotesize Note: Some percentages may not total 100 due to rounding.
\end{tablenotes}
\end{threeparttable}
\end{table}

Additionally, a separate set of 20 in-house radiology reports was used exclusively for prompt engineering to adapt SIFTING to the task of tumor staging.
These reports were not included in the annotation, evaluation, or statistical analyses.

The external validation set was annotated by a panel of three pulmonologists (NN, JM, and ML) and a thoracic radiologist (AN) with special interest in lung cancer and at least ten years of experience in diagnosing and managing patients with NSCLC ("truthers").
Annotations followed the protocol described by the 8th edition of the TNM Classification of Malignant Tumors \cite{brierley_j_tnm_2017}, which defines T-stage classification according to tumor size, local invasion into adjacent structures, and presence of satellite lesions.
Each truther assigned a single T-stage per report (T1a, T1b, T1c, T2a, T2b, T3, or T4), considering only findings that were positively confirmed.
For example, a nodule mentioned in the same lobe was not automatically treated as a satellite lesion unless explicitly stated, and statements such as "possible invasion" were not counted as positive evidence.
Reports lacking sufficient information for staging, or where staging was not applicable, were marked with an additional option indicating that the report is not suitable.
Full annotation instructions are provided in \ref{sec_annotation_instructions}.

The annotation process was conducted in two phases.
In Phase I (Blinded Phase), each truther reviewed all reports independently.
These initial annotations were used to assess interchangeability by comparing T-stage predictions to the annotations created by the individual truthers.
In Phase II (Unblinded Phase), the panel resolved discrepancies from Phase I through discussions to reach unanimous consensus for each report.
The consensus annotations, hereafter also referred to as the reference standard, served as the basis for evaluating the performance of the SIFTING framework and alternative approaches on the T-staging task.
The distribution of the T-stages in the final reference standard is reported in the Results section.

\subsection{SIFTING}
\subsubsection{Design}

SIFTING enables accurate and transparent information extraction from free-text, with all findings and values linked directly to the phrases from which they were derived, ensuring full traceability. 
A central feature to make this work is fine-grained control over the model output: by steering the token generation, the framework can implement reliable classification and generation tasks.
This control is achieved by constraining the model's token probabilities and guiding generation toward a restricted set of outcomes, a strategy commonly referred to as constrained decoding \cite{geng_grammar-constrained_2023, beurer-kellner_guiding_2024}. 
In classification tasks the model is presented with a set of predefined possible outcomes and is constrained to select one by only allowing the available option identifiers as valid responses.
For each option $o \in O$, the model computes a perplexity score, which estimates how likely the sequence of tokens representing a given option is given the prompt.
Formally, let $x_1^o, \dots, x_N^o$ be the tokens representing option $o$, and $P_\text{LM}(x_t^o \mid x_1^o, \dots, x_{t-1}^o)$ be the probability assigned by the language model to token $x_t^o$ given the preceding tokens.
The perplexity of the option is then:

\begin{equation}
\text{PPL}(o) = \exp\Bigg(
    -\frac{1}{N} \sum_{t=1}^{N} \log P_\text{LM}\big(x_t^o \mid x_1^o, \dots, x_{t-1}^o\big)\Bigg)
\end{equation}

The model selects the option with the lowest perplexity, ensuring that the choice aligns with the model's internal knowledge.
In generation tasks, the model can operate in two distinct modes.
In the first mode, it is constrained to quote exactly from the source text by using a deterministic finite automaton (DFA) that is created from all valid substrings of the text.
The DFA restricts the set of valid next tokens according to the current automaton state. 
At each step, only tokens continuing a valid path to an end-state in the DFA are allowed. 
In the second mode, the model is guided to produce output that adheres to a specified format, such as a regular expression (regex). 
For token-level control, the regex is also compiled into a DFA.
At each generation step $t$, let $C_t \subseteq V$ be the restricted subset of $V$ of allowed tokens determined by the DFA.
We define the constrained probability distribution as: 

\begin{align}
P_{C_t}(x_t \mid x_{1:t-1}) &=
\begin{cases}
\dfrac{P_\text{LM}(x_t \mid x_{1:t-1})}{Z_t}, & x_t \in C_t \\
0, & x_t \notin C_t
\end{cases} \\
\intertext{where}
Z_t &= \sum_{x' \in C_t} P_\text{LM}(x' \mid x_{1:t-1})
\end{align}

Here, $P_\text{LM}(x_t \mid x_{1:t-1})$ denotes the probability assigned by the underlying language model to token $x_t$ given the preceding context $x_{1:t-1}$ and $Z_t$ is the normalization constant (partition function) ensuring that $P_{C_t}$ sums to one over $C_t$.

\begin{figure}[t]
\centering
\includegraphics[width=\columnwidth]{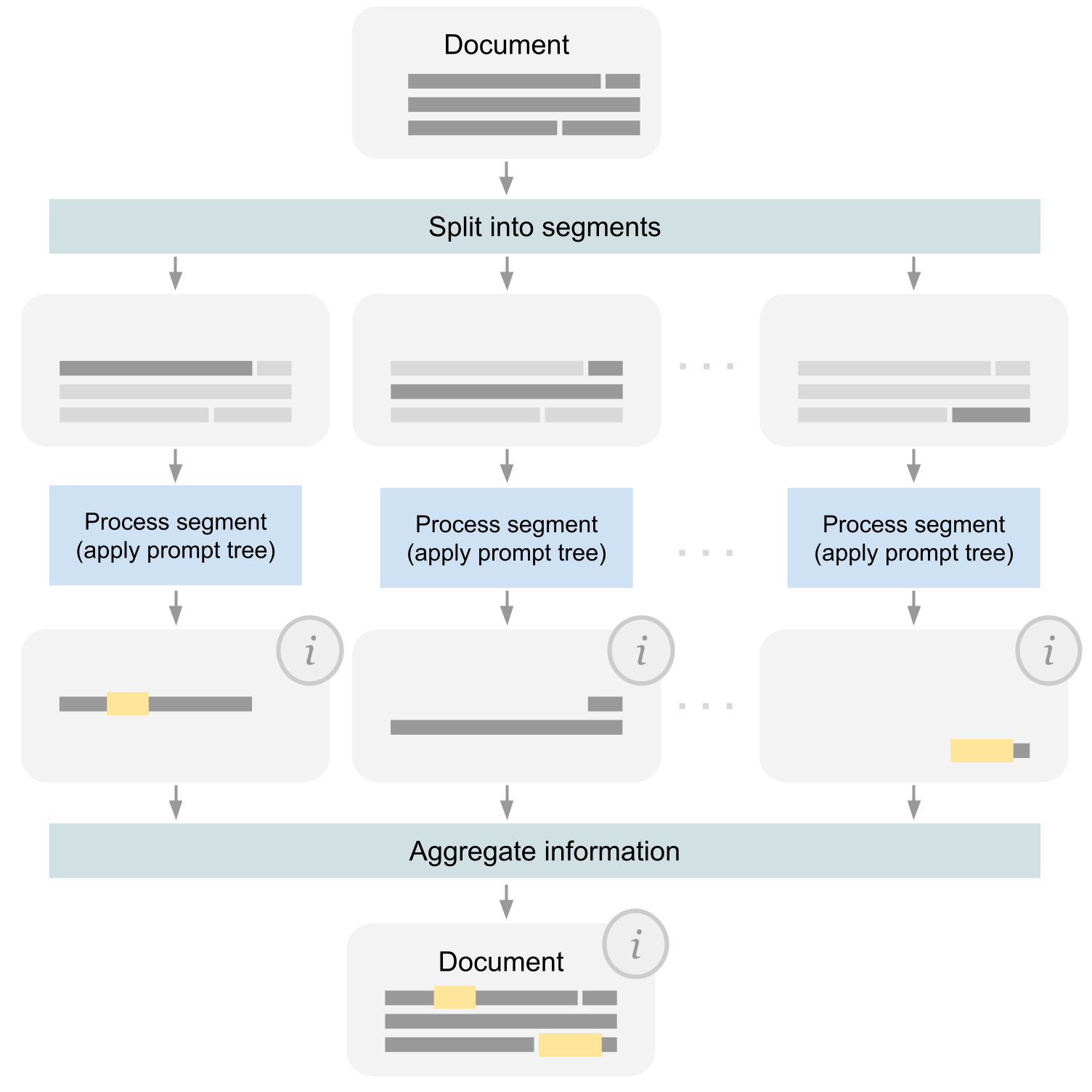}
\caption{Overview of the SIFTING processing flow. A document is processed segment by segment by applying a pre-defined series of classification and generation tasks on each.}
\label{fig_processing_global}
\end{figure}

Information extraction is performed by decomposing the complex document-level task of localizing information into a sequence of smaller, well-defined steps.
Documents are first segmented into sentence-level segments using a pre-existing text segmentation library \cite{frohmann_segment_2024}.
For each segment, surrounding text of configurable amount is included in the prompts to preserve contextual information.
Segments are then processed through an iterative series of classification and generation tasks to capture qualitative and quantitative information, respectively, following a prompt tree that captures all the attributes required to complete the desired task (see Figure \ref{fig_processing_global}).

\begin{figure*}[t]
\centering
\includegraphics[width=\textwidth]{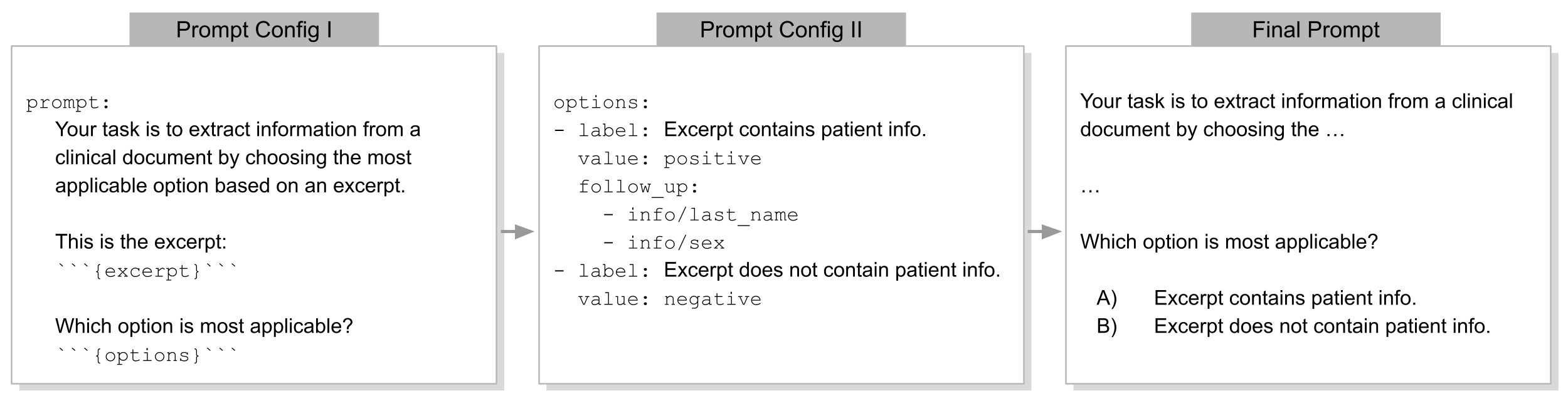}
\caption{Prompt tree example YAML files: the snippets show configuration files with placeholders and classification options and how they translate to the final prompt.}
\label{fig_prompting}
\end{figure*}

The prompt tree is defined through configuration and template YAML files, with placeholders that are dynamically filled at runtime.
Each prompt includes the target segment with context of configurable length, and task-specific instructions which may also contain few-shot examples.
YAML files are stored in a hierarchical directory layout which enables efficient management and extension of the prompt tree, with general task descriptions or instructions inherited from parent files and optionally overwritten at lower levels.
Classification task configurations define options with labels, values, and follow-up task identifiers.
Labels are automatically enumerated (e.g. \texttt{A}, \texttt{B}, \texttt{C}) and formatted into the prompt to complement the options.
Values record classification outcomes, while selected options trigger the corresponding follow-up tasks.
Generation task configurations instead specify explicit output constraints, such as quoting or adherence to a regex pattern.
Figure \ref{fig_prompting} provides an example of how prompts are constructed.

Performing many tasks for each segment is computationally demanding.
We make this approach feasible through two key technological features: (1) prompt caching, which avoids repeatedly reprocessing unchanged context, and (2) parallelization, which allows multiple segments and their applicable tasks to be processed simultaneously, optimizing resource utilization and overall efficiency.

\subsubsection{T-staging Task}
This section describes how the framework was applied to the task of determining the T-stage (hereafter referred to as SIFTING-T-stage) and provides an illustrative example of how qualitative and quantitative information is extracted from radiology reports and anchored to the relevant text segments. 

\begin{figure}[t]
\centering
\includegraphics[width=\columnwidth]{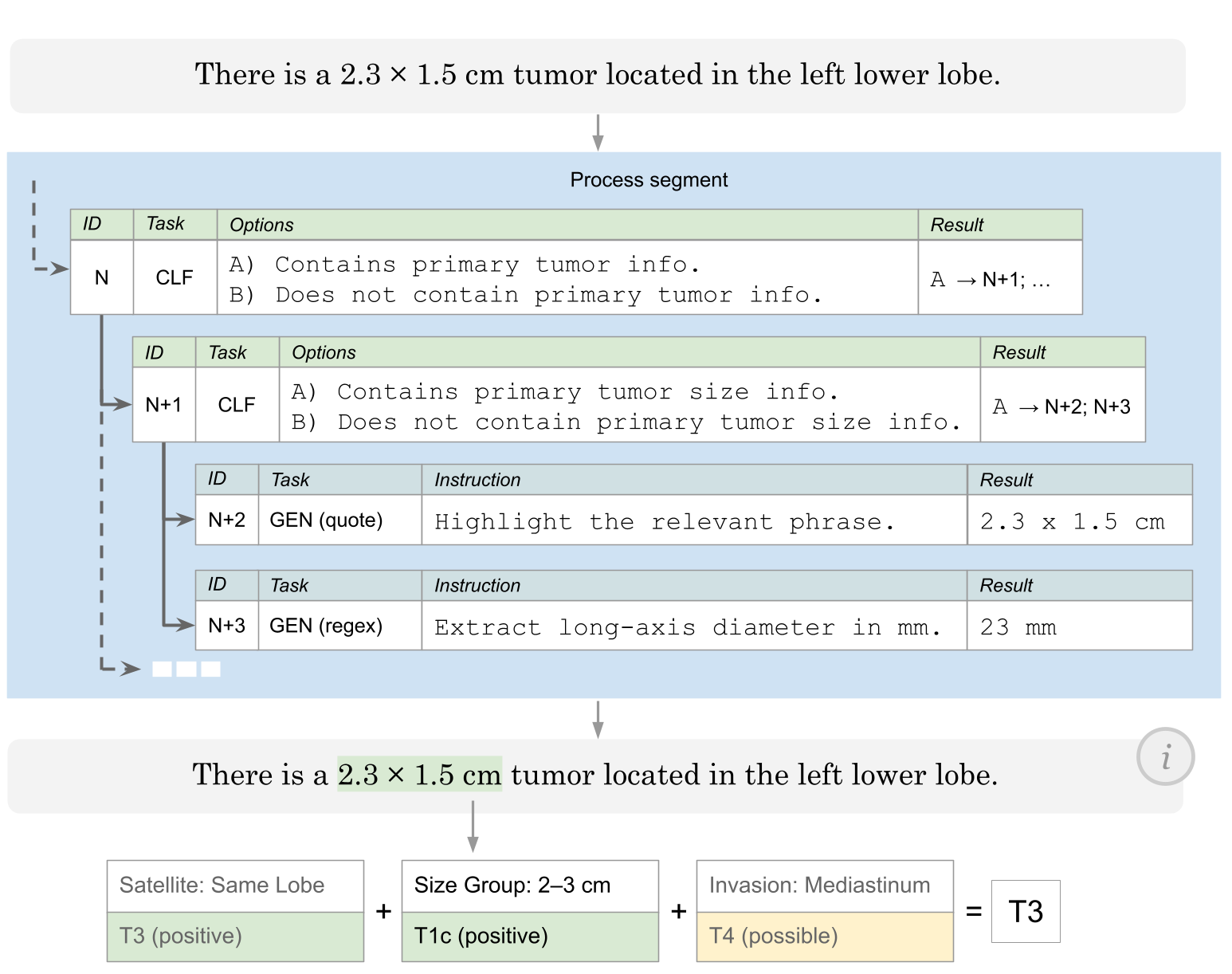}
\caption{Example of segment-level processing with SIFTING-T-stage.
The diagram illustrates the conditional application of classification and generation tasks to extract size information anchored to a specific phrase within the segment.
Once a segment containing relevant size information is identified, SIFTING-T-stage highlights the key phrase and simultaneously converts the size to millimeters, regardless of how it is presented in the report, using a simple regular expression (regex) pattern to constrain the model generation (i.e. \texttt{\textbackslash d\{1,3\}~mm}).
This approach ensures the size can be accurately assigned to the correct size group according to the staging protocol, enabling determination of T1c as the size-based T-stage.}
\label{fig_processing_segment}
\end{figure}

A prompt tree with classification and generation tasks was developed to closely follow the protocol as provided by the 8th edition of the TNM Classification of Malignant Tumors \cite{brierley_j_tnm_2017}, ensuring that all relevant information for T-staging is captured.
Classification tasks produce qualitative findings that are directly tied to specific segments.
Each finding is designed to be categorized as 'positive', 'possible', or 'negative'.
A positive finding indicates that the feature is clearly present (e.g. 'the tumor invades the pleura').
A possible finding suggests the presence of a feature but does not confirm it (e.g. 'the tumor may invade the pleura'), and a negative finding explicitly indicates its absence (e.g. 'no pleural invasion is observed').
In this study, only positive findings contributed to the determination of the overall T-stage.
Findings classified as possible or negative still got highlighted in the source text for clinician review, providing transparency and enabling expert verification, but they did not influence the final T-stage.
Generation tasks capture exact phrases and standardized quantitative values.
In this task, extracted quantitative size values are mapped to T-stage specific size groups.
Once all segments are processed, the extracted information is aggregated to determine the overall T-stage for the report.
The highest T-stage indicated by positive findings determines the overall stage, reflecting the most severe pathology present in the report. 
This segment-level processing flow is visualized in Figure \ref{fig_processing_segment}.

\subsubsection{Inference}
Inference was performed on a local server with 512~GB of RAM and a single NVIDIA L40S GPU with 48~GB of memory.
SIFTING-T-stage uses a compact 4-bit quantized version of Meta's Llama-3.3-70B with a resulting size of 35~GB.
Report segments are processed in batches and in parallel, applying the classification and generation tasks defined in the prompt tree.
Caching was employed to improve efficiency and reduce redundant computation by reusing previously processed context and prompt components where possible.
The key-value (KV) cache was partially offloaded to CPU memory to allow longer context windows.

\subsection{Evaluation and Statistical Analysis}
All analyses were conducted retrospectively using previously annotated reports and inference results. Performance metrics and visualizations are provided to illustrate model behavior and agreement with clinical experts.

\subsubsection{Assessment of Interchangeability with Truthers}
A bootstrap-based analysis was performed to assess whether SIFTING-T-stage could be considered non-inferior to human readers.
The analysis was based on Phase I annotations, where each truther independently reviewed all reports, preserving natural variability.
For each bootstrap resample (sampling reports with replacement), one truther was randomly replaced with SIFTING-T-stage, and inter-rater agreement was calculated for both the resulting hybrid panel and the original all-human panel, yielding paired agreement scores.
Across 5,000 bootstrap samples, the paired differences were summarized, and 95\% percentile bootstrap confidence intervals were computed.
To formally evaluate comparability, we conducted a one-sided non-inferiority test with a margin of 0.5\%, testing whether the hybrid panel’s agreement was non-inferior to that of the all-human panel.
For reference, point estimates of the agreement scores were calculated for all pairwise rater comparisons using the original dataset.

\subsubsection{Model Performance Assessment}
The overall performance of SIFTING-T-stage and an alternative single-prompt approach was evaluated against the Phase II consensus annotations.
Variability and uncertainty were quantified using bootstrap resampling of cases, yielding distributions of overall T-stage prediction accuracy.
The alternative approach used a single prompt requesting both a T-stage prediction and a justification, with a comprehensive staging summary and instructions analogous to those given to the truthers (full prompt in Appendix \ref{fig_global_prompt}).
This approach was executed across multiple models varying in size, openness (open vs. closed source), and hosting platform, each producing a full set of predictions for comparison.
In the Results chapter, we report model-specific accuracies and compare their distributions with those of SIFTING-T-stage.
Two-sided statistical tests were used to assess whether observed differences in accuracy were significant.

\section{Results}
\begin{figure}[t]
\centering
\includegraphics[width=\columnwidth]{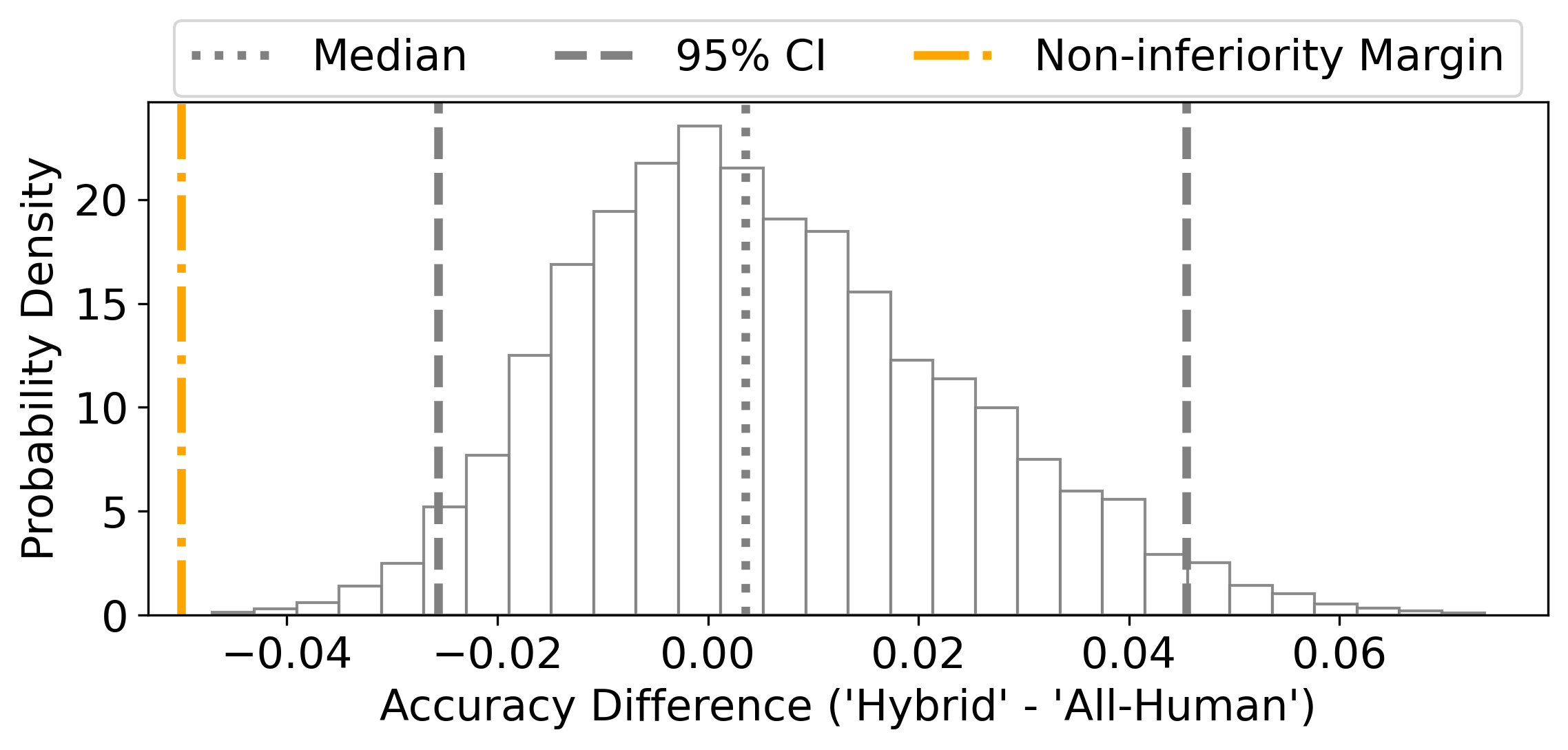}
\caption{Distribution of bootstrap difference scores (hybrid panel – all-human panel) in the interchangeability analysis, shown as a normalized histogram representing a probability distribution.
The figure shows the empirical distribution of differences, together with the median, the 95\% confidence interval, and the non-inferiority margin of 0.5\%. 
With the 2.5th percentile of the bootstrap distribution exceeding the margin, the analysis demonstrates statistical non-inferiority of the hybrid panel relative to the all-human panel.}
\label{fig_interchangeability}
\end{figure}

SIFTING-T-stage was assessed for interchangeability with human truthers.
Across 5,000 bootstrap resamples, the hybrid panel, where one truther was replaced by SIFTING-T-stage per resample, achieved a mean pairwise agreement of 82.2\% (SD = 2.8\%), while the all-human panel achieved a mean pairwise agreement of 81.6\% (SD = 2.5\%).
Using a 0.5\% non-inferiority margin, the one-sided significance test confirmed that the hybrid panel's agreement was statistically non-inferior to that of the all-human panel (p $<$ 0.001).
The distribution of paired differences is shown in Figure \ref{fig_interchangeability}.
To complement this analysis, we calculated the pairwise agreement between SIFTING-T-stage and each individual truther, as well as between each truther pair, on the original dataset.
Point estimates are reported in Table \ref{table_pairwise}.
Of 130 total reports, the number deemed stageable was 121 for SIFTING-T-stage, and 117, 121, 117, and 119 for Truthers A, B, C, and D, respectively.

\begin{table}[htbp]
\centering
\begin{threeparttable}
\small
\caption{Pairwise Agreement.}
\label{table_pairwise}
\begin{tabularx}{\columnwidth}{X c@{\hskip 7pt}c@{\hskip 7pt}c@{\hskip 7pt}c}
\toprule
 & Truther A & Truther B & Truther C & Truther D \\
\midrule
SIFTING-T-stage & 0.83 & 0.85 & 0.84 & 0.78 \\
Truther A       &      & 0.87 & 0.82 & 0.81 \\
Truther B       &      &      & 0.81 & 0.82 \\
Truther C       &      &      &      & 0.77 \\
\bottomrule
\end{tabularx}
\end{threeparttable}
\end{table}

\begin{figure}[htbp]
    \centering
    \begin{subfigure}[t]{0.49\columnwidth}
        \centering
        \includegraphics[width=\linewidth]{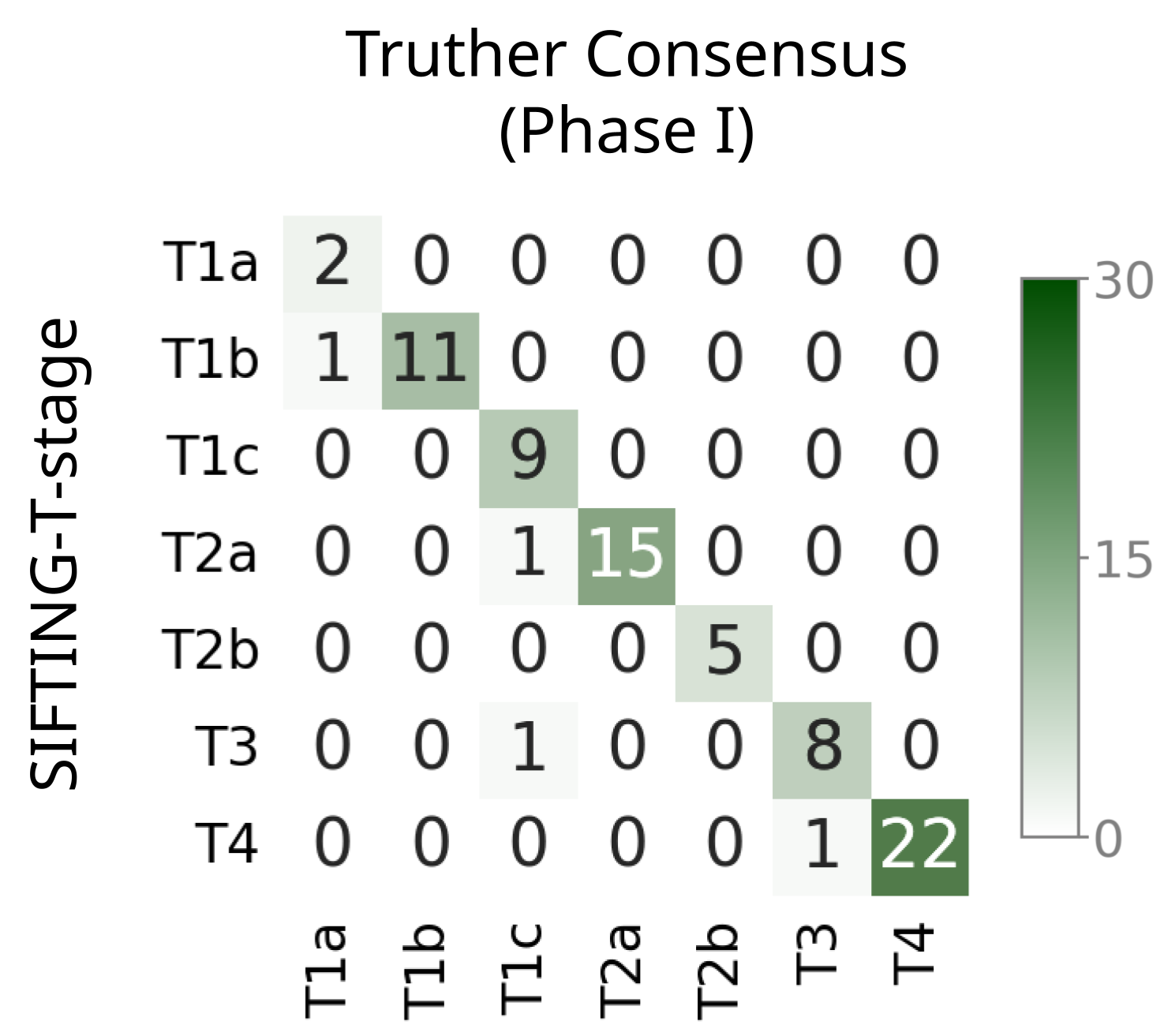}
        \caption{}
    \end{subfigure}
    \hfill
    \begin{subfigure}[t]{0.49\columnwidth}
        \centering
        \includegraphics[width=\linewidth]{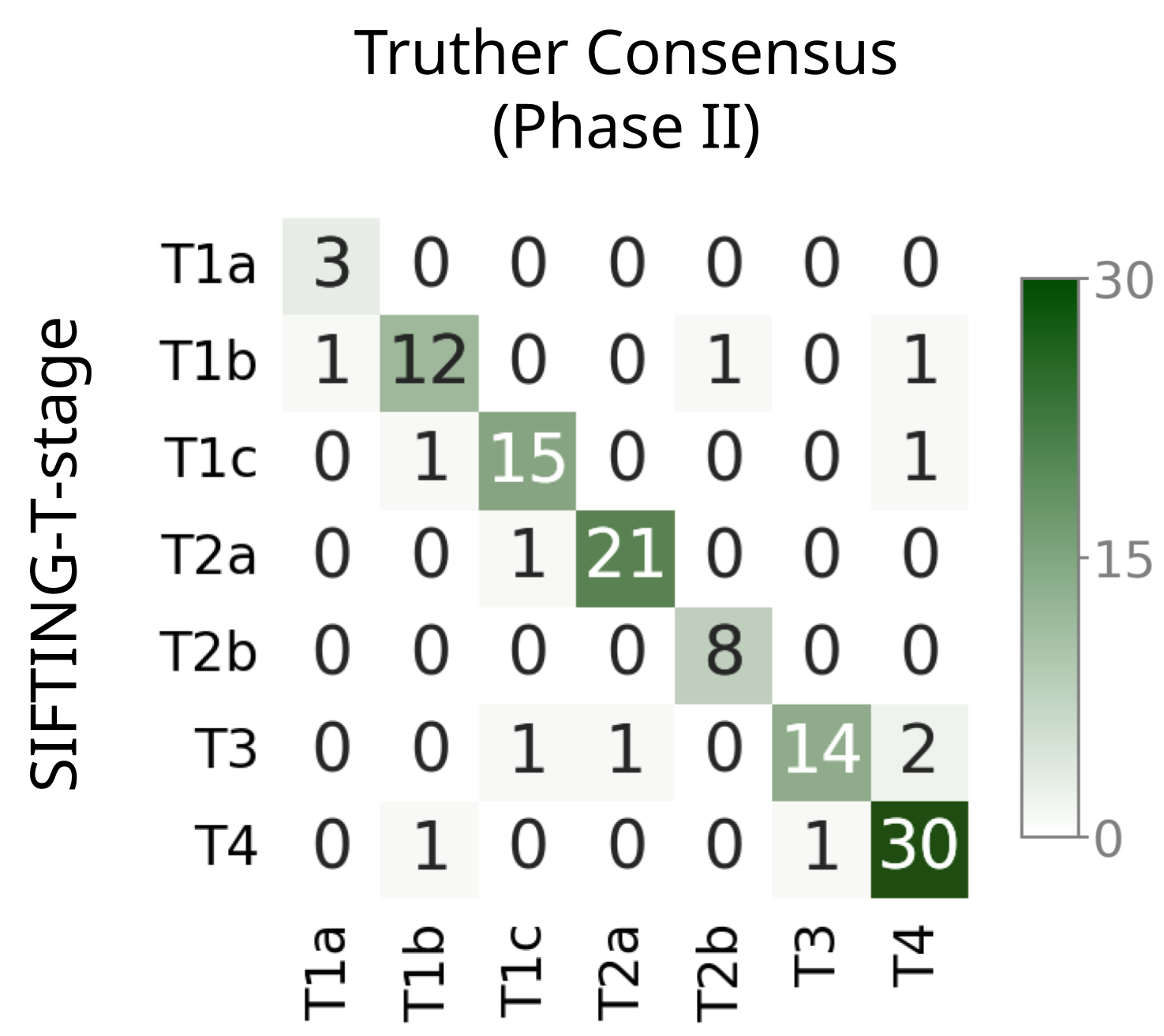}
        \caption{}
    \end{subfigure}
    \caption{Confusion matrices comparing SIFTING-T-stage predictions with T-stages unanimously agreed upon by four human truthers. Left: Phase~I (Blinded Phase); Right: Phase~II (Unblinded Phase). 
}
    \label{fig_confusion}
\end{figure}

The confusion matrices in Figure \ref{fig_confusion} compare SIFTING-T-stage predictions with T-stages that have been agreed upon unanimously by all four truthers, established after Phases I and II.
After Phase I, 77 reports were unanimously agreed upon by the truthers, yielding 76 valid pairs for comparison with SIFTING-T-stage; after Phase II, 118 reports unanimously agreed upon, yielding 115 valid pairs.
Most classifications lie along the diagonal, indicating strong agreement with the consensus annotations.
SIFTING-T-stage made 4 misclassifications based on the Phase I data (all over-staged) and 12 based on the Phase II data (7 over-staged, 5 under-staged).
An error analysis of the 12 misclassifications against the Phase II data, i.e. the reference standard, is presented in Table \ref{table_mistakes}, grouping errors by type and providing brief descriptions.

\newcolumntype{L}[1]{>{\raggedright\arraybackslash}p{#1}} 
\newcolumntype{Y}{>{\raggedright\arraybackslash}X}        

\begin{table*}[htbp]
\centering
\begin{threeparttable}
\caption{Error analysis of SIFTING-T-stage outputs.}
\label{table_mistakes}
\renewcommand{\arraystretch}{1.5}
\small
\begin{tabularx}{\textwidth}{L{3cm} L{3cm} Y c}
\toprule
Error Group & Error Type & Description & \# \\
\midrule
Missed confirmation & Invasion & Report describes invasion with sufficient certainty, but SIFTING-T-stage did not mark as confirmed invasion. & 1 \\
 & Satellite nodules & Report describes ipsilateral satellite nodules with sufficient certainty (“diffuse metastases in all lobes of the lung”), SIFTING-T-stage did not mark as confirmed satellite nodules. & 1 \\
\addlinespace
Ambiguity & Lesion origin & When multiple nodules were present, SIFTING-T-stage selected a different primary tumor nodule than the thruthers. & 6 \\
\addlinespace
Instruction adherence & Satellite nodules & SIFTING-T-stage followed instructions to avoid automatically assigning satellite status to additional nodules; truthers considered context in interpreting likely satellite nodules. & 2 \\
 & Primary tumor & SIFTING-T-stage selected a nodule as primary tumor as per instructions; truthers applied domain knowledge to include or exclude nodules based on descriptors indicating malignancy or benignness, respectively. & 2 \\
\bottomrule
\end{tabularx}
\end{threeparttable}
\end{table*}

\begin{figure}[htbp]
    \centering
    \begin{subfigure}[t]{0.9\columnwidth}
        \centering
        \includegraphics[width=\linewidth]{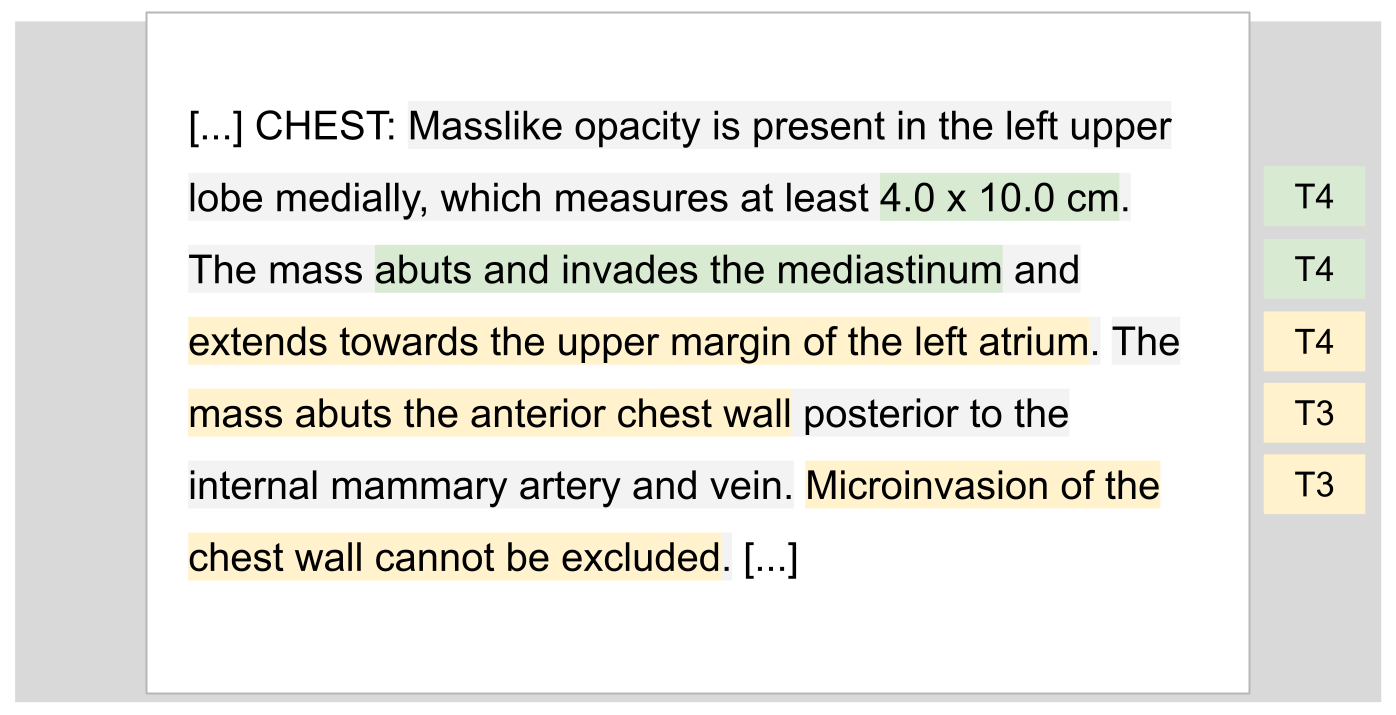}
        \caption{}
    \end{subfigure}
    
    \vspace{0.5em} 

    \begin{subfigure}[t]{0.9\columnwidth}
        \centering
        \includegraphics[width=\linewidth]{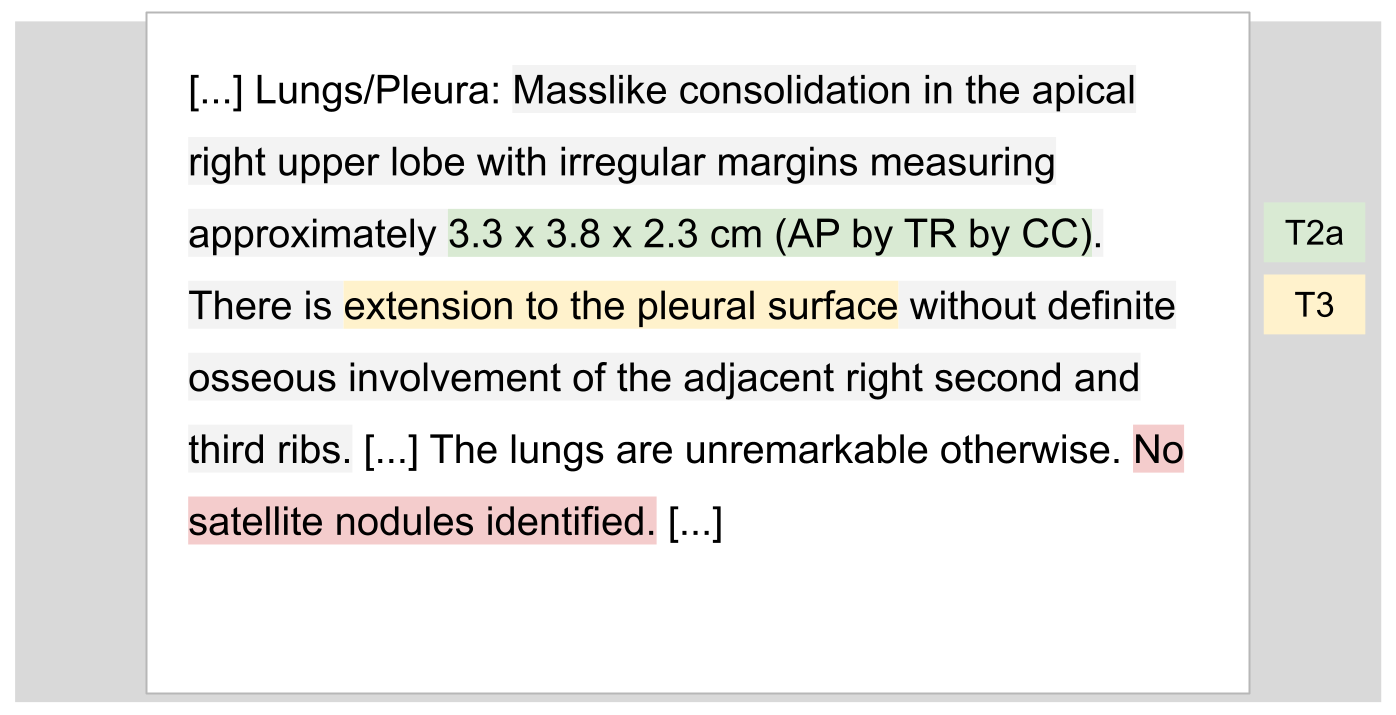}
        \caption{}
    \end{subfigure}
    
    %

    \caption{
Radiology report extracts highlighting tumor staging information.
Segments containing relevant information are shaded in gray. Within each segment, phrases used as anchors for extracted qualitative or quantitative information are marked in green, yellow, or red, indicating positive, possible, or negative findings, respectively. Example (a) shows two confirmed (positive) and three unconfirmed (possible) findings. Example (b) includes a negative finding and illustrates a possible upstage from T2a to T3.}
    \label{fig_highlights}
\end{figure}

To illustrate how extracted information is anchored to the source text, we present real examples of visual text highlights as provided by SIFTING-T-stage in Figure \ref{fig_highlights}.

SIFTING-T-stage achieved an overall accuracy of 90\% across all T-stages against the reference standard, and 91\% when sub-stages were aggregated into the main stages T1–T4. 
Tables \ref{table_report_full} and \ref{table_report_aggregated} report precision, recall, and F1-score for each stage, along with the number of reports per stage (support), for all individual T-stages and for the main stages, respectively.

\begin{figure*}[t]
\centering
\includegraphics[width=0.9\textwidth]{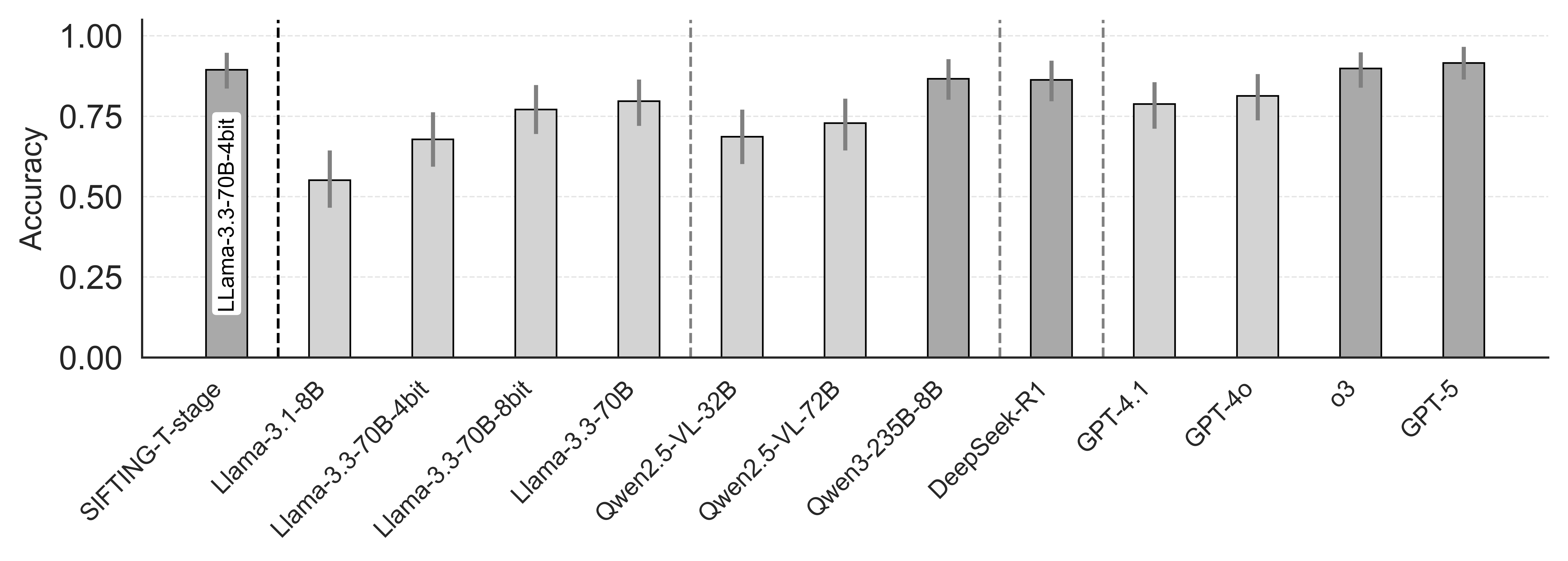}
\caption{
Overall T-stage accuracy for SIFTING-T-stage compared with the alternative single-prompt approach across a range of LLMs. 
Bars show bootstrap distributions (5,000 resamples) with 95\% confidence intervals. 
Dashed vertical lines separate SIFTING-T-stage from the alternative models, and further group models by manufacturer for easier visual interpretation.
The label “Llama-3.3-70B-4bit” inside the SIFTING-T-stage bar indicates that this is the LLM that was utilized by the framework.
Models statistically indistinguishable from SIFTING-T-stage (p $\geq$ 0.05) are shaded dark gray.
}
\label{fig_barplot}
\end{figure*}

Figure \ref{fig_barplot} compares the performance of SIFTING-T-stage with the performance of a range of LLMs as used in the alternative single-prompt approach. 
Each bar represents the distribution of the T-stage prediction accuracy based on 5,000 bootstrap resamples of the reports, allowing direct visual comparison.
Notably, the 4-bit quantized version of Llama-3.3 70B, when used in the single-prompt approach, achieved an accuracy of only 68\%, while when used within SIFTING-T-stage, an accuracy of 90\% was achieved.
Table \ref{table_comparison} provides all the mean accuracy scores along with the 95\% confidence intervals, as well as p-values to indicate which models are statistically different from SIFTING-T-stage based on the accuracy distributions over the bootstrap resamples.
The models that are not statistically different from SIFTING-T-stage based on these results are also marked in Figure \ref{fig_barplot} through darker shading.
Table \ref{table_comparison} further provides information about model metadata including manufacturer, release year, open-source status, and an approximate size estimate where possible.

\begin{table}[htbp]
\centering
\begin{threeparttable}
\small
\caption{Classification results -- all T-stages}
\label{table_report_full}
\begin{tabularx}{\columnwidth}{X r r r r}  
\toprule
 & Precision & Recall & F1-score & Support \\
\midrule
T1a          & 1.00 & 0.75 & 0.86 & 4  \\
T1b          & 0.80 & 0.86 & 0.83 & 14 \\
T1c          & 0.88 & 0.88 & 0.88 & 17 \\
T2a          & 0.95 & 0.95 & 0.95 & 22 \\
T2b          & 1.00 & 0.89 & 0.94 & 9  \\
T3           & 0.78 & 0.93 & 0.85 & 15 \\
T4           & 0.94 & 0.88 & 0.91 & 34 \\
\midrule
Weighted Average & 0.90 & 0.90 & 0.90 & -- \\
\bottomrule
\end{tabularx}
\end{threeparttable}
\end{table}

\begin{table}[htbp]
\centering
\begin{threeparttable}
\caption{Classification results -- aggregated into main T-stages}
\label{table_report_aggregated}
\small
\begin{tabularx}{\columnwidth}{X r r r r}  
\toprule
 & Precision & Recall & F1-score & Support \\
\midrule
T1              & 0.91 & 0.91 & 0.91 & 35 \\
T2              & 0.97 & 0.94 & 0.95 & 31 \\
T3              & 0.78 & 0.93 & 0.85 & 15 \\
T4              & 0.94 & 0.88 & 0.91 & 34 \\
\midrule
Weighted Average & 0.92 & 0.91 & 0.91 & -- \\
\bottomrule
\end{tabularx}
\end{threeparttable}
\end{table}


\begin{table*}[htbp]
\centering
\begin{adjustbox}{max width=1.3\textwidth, center}
\begin{threeparttable}
\caption{Performance of SIFTING-T-stage and LLMs with model info}
\label{table_comparison}
\small
\begin{tabular}{l r r r@{\hspace{0.2em}}l r r c c r}
\toprule
 & \multicolumn{5}{c}{\textbf{Performance}} & \multicolumn{4}{c}{\textbf{Model Info}} \\
\cmidrule(lr){2-6} \cmidrule(lr){7-10}
Model & Mean Accuracy & 95\% CI & $p$-val & & Accepted (\%) & Manufacturer & Released & Open Source & Size (GB) \\
\midrule
SIFTING-T-stage    & 0.90 & 0.84--0.95 & --       &      &115 (97.5)  & --   & --   & --  & --  \\
\midrule
Llama-3.1-8B       & 0.55 & 0.47--0.64 & $<$0.001 & {*}  & 118 (100.0) & Meta     & 2024 & Yes & 16  \\
Llama-3.3-70B-4bit & 0.68 & 0.59--0.76 & $<$0.001 & {*}  & 118 (100.0) & Meta     & 2024 & Yes & 35  \\
Llama-3.3-70B-8bit & 0.77 & 0.69--0.85 & 0.006    & {*}  & 118 (100.0) & Meta     & 2024 & Yes & 70  \\
Llama-3.3-70B      & 0.80 & 0.72--0.86 & 0.026    & {*}  & 118 (100.0) & Meta     & 2024 & Yes & 140 \\
Qwen2.5-VL-32B     & 0.69 & 0.60--0.77 & $<$0.001 & {*}  & 118 (100.0) & Alibaba  & 2024 & Yes & 64  \\
Qwen2.5-VL-72B     & 0.73 & 0.64--0.81 & $<$0.001 & {*}  & 118 (100.0) & Alibaba  & 2024 & Yes & 144 \\
Qwen3-235B-8B      & 0.87 & 0.80--0.93 & 0.387    &      & 113 (95.8)  & Alibaba  & 2025 & Yes & 235 \\
DeepSeek-R1        & 0.86 & 0.80--0.92 & 0.408    &      & 117 (99.2)  & DeepSeek & 2025 & Yes & 720  \\
GPT-4.1            & 0.79 & 0.71--0.86 & 0.006    & {*}  & 118 (100.0) & OpenAI   & 2023 & No  & --  \\
GPT-4o             & 0.81 & 0.74--0.88 & 0.048    & {*}  & 118 (100.0) & OpenAI   & 2024 & No  & --  \\
o3                 & 0.90 & 0.84--0.95 & 0.894    &      & 118 (100.0) & OpenAI   & 2025 & No  & --  \\
GPT-5              & 0.92 & 0.86--0.97 & 0.503    &      & 118 (100.0) & OpenAI   & 2025 & No  & --  \\
\bottomrule
\end{tabular}
\begin{tablenotes}[flushleft]
\item{Note: p-values indicate results of bootstrap-based two-sided tests comparing each LLM's performance with SIFTING-T-stage.}
\end{tablenotes}
\end{threeparttable}
\end{adjustbox}
\end{table*}

\section{Discussion}

In this work we introduce SIFTING, a novel framework for gathering information from clinical free-text with full traceability.
Recent reviews highlight that automated tools in healthcare must provide outputs that are stable, reproducible, and auditable, with clear links to source data to ensure compliance and patient safety \cite{huang_explainable_2024}.
While classical NLP methods are more inherently transparent, they lack the adaptability of modern LLMs \cite{mcnicholas_natural_2025}.
To bridge this gap, SIFTING combines the strengths of LLMs with segment-by-segment processing and fine-grained output control, enabling interpretability and traceability beyond current approaches.
Well-defined classification and generation tasks limit verbosity and constrain the model to produce only structured, relevant information, thereby effectively reducing the risk of hallucinations by design.
We demonstrate its utility by applying the framework to the task of extracting tumor staging information from radiology reports (SIFTING-T-stage).

Our results show that agreement of the SIFTING-T-stage with individual truthers was comparable to the agreement observed among truthers themselves.
Non-inferiority testing based on bootstrapped resamples, in which individual truthers were randomly substituted, confirmed that SIFTING-T-stage was statistically non-inferior (p $<$ 0.001).
These findings suggest that SIFTING-T-stage may be considered interchangeable with a human truther in accomplishing this task in this setup.
On the final reference standard, the framework achieved a high point estimate accuracy of 90\% across the suitable reports.
Notably, this performance was achieved without model training or parameter tuning, relying solely on a carefully designed prompt tree and instructions derived from a small set of representative radiology reports.
This highlights the potential of LLM-based frameworks to perform complex clinical information extraction with minimal training data requirements.


For benchmarking, an alternative single-prompt approach was evaluated against the reference standard.
This approach used a prompt containing the staging system summary and brief instructions on handling uncertain findings (see Figure \ref{fig_global_prompt} in the appendix).
Several small and large LLMs were tested (see Figure \ref{fig_barplot}), producing point estimate accuracy scores ranging from 55\% to 92\%.
Statistical testing indicated no significant difference in performance between SIFTING-T-stage and the best-performing models in the single-prompt approach.
This demonstrates that SIFTING-T-stage can achieve accuracy scores comparable to state-of-the-art LLMs with expensive reasoning capabilities, while providing structured, traceable outputs tailored for use in clinical practice. 

The results obtained by SIFTING-T-stage and the alternative single-prompt approach are comparable to those presented in previous related works.
Notably, Nobel et al. addressed an almost identical T-staging task using conventional NLP methods.
Their system achieved similarly high accuracy scores. However, they also describe a substantial manual effort that was required to capture terminology, synonyms, and contextual rules \cite{nobel_natural_2020, nobel_t-staging_2021}.
Other more recent studies have approached staging tasks with LLM-based approaches with promising results \cite{matsuo_exploring_2024, fink_potential_2023}.
As an example, Lee et al. applied an approach similar to our single-prompt approach to the broader task of complete lung cancer staging \cite{lee_lung_2024}.
Their best model (GPT-4o) achieved an overall accuracy of 74.1\%.
While still below fellowship-trained radiologists, this performance was comparable to clinical fellows and required minimal manual intervention. Together, our results and prior work demonstrate the power and potential of modern SOTA models.
However, imperfect accuracy, combined with a lack of interpretability and traceability, suggest that off-the-shelf LLMs may not yet be fit to complete complex higher-level clinical tasks autonomously \cite{hager_evaluation_2024, busch_current_2025, ullah_challenges_2024}.

SIFTING addresses this gap by embedding traceability directly into the process, matching or surpassing performance scores from previous studies while providing the interpretability required for reliable clinical use.
Traceable text source highlights, as created by SIFTING-T-stage and illustrated in Figure \ref{fig_highlights}, have the potential to support clinician review and discussion of ambiguous cases.

Radiology reports are known to contain ambiguities and require interpretation, introducing variability in how information is understood \cite{lukaszewicz_art_2016}.
To enable an objective and unified approach to T-staging in the context of this study, we provided truthers with a set of explicit instructions on how to handle ambiguities (see \ref{sec_annotation_instructions}).
Not being able to consult the imaging, and having to make certain assumptions, makes the task of this study somewhat different from that in clinical practice.
The task is well suited to demonstrate the capabilities of SIFTING, including its ability to highlight relevant sections or areas needing clarification.
However, due to the differences with standard clinical practice, strong performance on this task and on the ones presented in previous research, does not imply that automated information extraction tools can readily replace expert clinicians.
Rather, we see the potential of frameworks such as SIFTING-T-stage to complement clinical workflows by supporting interpretation and review. 

In our study, discussion was needed to reach consensus on a total of 53 cases, highlighting that interpretative differences and ambiguities were relatively common in this dataset.
This suggests that the final reference standard may retain some degree of subjectivity, making perfect agreement by any model inherently challenging.
Consistent with this, Table \ref{table_mistakes} shows that most model errors (6 in total) arose from ambiguity in determining which nodule should be classified as the intrapulmonary primary tumor.
These findings underscore the value of traceable and transparent outputs, as visualizing relevant report sections can facilitate discussion and support clinical decision-making.

Adhering to contemporary guidelines and standards when creating radiology reports is a critical step toward producing clear, unambiguous reports and improving clinical workflows \cite{gershanik_critical_2011}.
Standardized structure and terminology enhance communication with referring physicians and improve the performance of automated systems for data extraction and analysis \cite{vosshenrich_investigating_2023}.
Avoiding excessive synonyms and loosely phrased sentences, and using one idea per sentence with consistent terminology, further improves the accuracy and reliability of automated extraction \cite{leaman_challenges_2015}.
In conclusion, these results suggest that frameworks like SIFTING with traceability are especially valuable in contexts where document content is inherently variable or partially ambiguous.

With SIFTING, each extracted piece of information is directly linked to the exact phrase from which it was obtained.
Information extraction is performed by decomposing the document-level task into a sequence of smaller, isolated steps. 
Each document is processed segment by segment using an iterative prompt tree that mirrors the logical structure of the information to be extracted.
Rather than trying to capture the full complexity of medical language, the challenge shifts to creating prompts with clear task instructions. 
Narrow, well-defined tasks allow more efficient use of a model’s capacity, and combined with precise output control, even relatively small models can achieve high performance.
For example, as demonstrated in this study and visualized in Figure 6, using the 4-bit quantized version of Llama-3.3-70B with a size of only 35~GB, SIFTING-T-stage achieved a point estimate accuracy of 90\% on the reference standard.
In contrast, when the same model was utilized in the alternative single-prompt approach, accuracy dropped to 68\%.
Models are required to follow instructions closely, but smaller models may struggle when instructions conflict with or challenge their internal knowledge \cite{murthy_evaluating_2024}.
Consequently, approaches like SIFTING are still likely to benefit further from future model improvements.

Closed models such as GPT-4 or Claude do not provide the level of access required for strict output control.
Their behavior may vary across versions, outputs can be non-deterministic, and hallucinations or paraphrasing cannot always be prevented, limiting traceability and reproducibility \cite{atil_non-determinism_2025}.
Externally hosted open-source models typically lack token-level hooks necessary for fine-grained output control.
However, full access to the model generation process is required to implement methods like SIFTING and meet the reliability and transparency standards needed in regulated domains.
Hosting large open-source models locally is operationally demanding.
Models in the 70B–100B+ parameter range often require multi-GPU setups, high memory bandwidth, and careful orchestration.
Smaller models, such as 7B parameters, are easier to run but might lack sufficient capacity for reliable performance, even on the focused tasks generated by SIFTING.
Models in the 70B parameter class offer a practical balance: they are capable of robust instruction-following, handle messy clinical input consistently, and allow fine-grained token-level control.
With 4-bit quantization and optimized memory management, these models can be deployed efficiently on a single 48~GB GPU without compromising performance too much.

Processing documents segment by segment, with multiple classification and generation tasks per segment, increases the total number of tasks and, thus, inference time.
SIFTING alleviates this computational burden by employing parallelization and caching.
In this study, the number of tasks per report had a median of 126 (IQR: 50–213), resulting in a mean inference time of 2.3 minutes across the 130 reports on a single NVIDIA L40S GPU (48 GB VRAM) with parallel processing and caching (per-case variability not defined under parallel execution), compared to 3.6 minutes (SD = 1.4) without these optimizations.

Effective prompt design is central to SIFTING.
Prompts are guided primarily by clinical guidelines or downstream data requirements, and can include instructions or few-shot examples to illustrate nuances, such as how to handle hedging language.
For the T-staging task, a small set of sample reports was enough to create the prompts, which capture the guideline’s structured logic rather than the full variability of report language.
The prompts are arranged hierarchically in a prompt tree, enabling stepwise, traceable information extraction.
Manual construction of prompt trees ensures accuracy and completeness, though it may demand significant effort and domain knowledge depending on the task.
Future work could explore semi-automated or agentic approaches to streamline this process, potentially further reducing manual effort while maintaining reliability.

SIFTING has the potential to support a variety of downstream tasks, like populating standardized forms with patient-level data, such as demographics, lab results, prior treatments, or specialized information like biomarker status from Next-Generation Sequencing (NGS) reports.
Its controlled generation and classification capabilities, with predefined options and strict output constraints, ensure that extracted data adheres to required formats, supporting compliance with backend database and form validation checks.
Future work could explore additional applications across different clinical domains and integration with electronic health record systems, leveraging SIFTING’s transparent extraction capabilities to enhance clinical workflows.

\section{Conclusion}

LLMs are promising for extracting information from clinical free-text, but their outputs often lack traceability.
By combining LLMs with segment-level processing and structured prompts with strict output control, SIFTING represents a framework for accurate and transparent information extraction.
Demonstrated on tumor T-stage information extraction from radiology reports, it achieved high performance without requiring additional model training or large labeled datasets.
The framework's traceable, structured outputs can reduce manual effort, support clinical decision-making, and be extended to other tasks requiring reliable information extraction from clinical documents.

\section*{Declaration of generative AI and AI-assisted technologies in the manuscript preparation process}
During the preparation of this work, the authors used ChatGPT (\mbox{OpenAI}) to assist in the writing process to improve the readability and language of the manuscript. After using this tool, the authors reviewed and edited the content as needed and take full responsibility for the content of this work.

\section*{Ethics Statement}
This study did not require institutional ethics approval as it used only de-identified, retrospective data obtained from a commercial data provider.
The data provider confirmed that all source data were collected in compliance with applicable laws and regulations, including informed consent and privacy protection requirements.
No direct interaction with human subjects occurred, and no identifiable personal information was accessed or processed by the authors.

\section*{Data Statement}
The data used in this study are proprietary and were obtained under a commercial license.
They cannot be shared publicly due to licensing and confidentiality restrictions.

\appendix

\section{Annotation instructions}
\label{sec_annotation_instructions}
The following specific annotation instructions have been given to the truthers to accompany the staging guidelines as provided by the 8th edition of the TNM Classification of Malignant Tumors \cite{brierley_j_tnm_2017}:

\vspace{1em}
\noindent\rule{\linewidth}{1pt} 
\paragraph{Assumptions} You may assume the following:
\begin{itemize}
    \item[--]{The patient has a confirmed lung cancer diagnosis.}
    \item[--]{The largest intrapulmonary lesion described in the report is assumed to be the primary tumor for staging.}
\end{itemize}

\paragraph{General Guidelines}
Only use explicitly confirmed information for staging. Do not infer or assume findings beyond what is clearly stated in the report. T staging should be based solely on confirmed details regarding tumor size, invasion, or spread:

\begin{itemize}
    \item[--] If a process is described with phrases that convey uncertainty such as "may invade", "possibly extends into" or "suggestive of", it must not be considered for T-staging unless definitive invasion is also explicitly confirmed.

    \item[--] If a process is described with terms that only indicate proximity such as "abuts", "occludes", "encases," "indents," or "surrounds", it must not be considered for T-staging unless definitive invasion is also explicitly confirmed.

    \item[--] Do not assume that any additional nodule is automatically a satellite nodule. 
    Only classify separate nodules as satellite nodules if they are described as such or if their relationship to the primary tumor is explicitly stated or strongly implied.
\end{itemize}

We are interested in the current size of the largest intrapulmonary nodule, mass, or tumor that is described by the report. Consider the largest dimension if multiple dimensions are provided.
If a dimension is given without a unit, but another dimension (e.g.  anterior-posterior, transverse, cranio-caudal) in the same measurement has a unit (e.g. cm), it is assumed that the unit applies to all dimensions unless otherwise specified.
\\
\\
Determine whether it is possible to provide a T-stage using the information given in the radiology report.
Your label options are: yes / no
\\
\\
Determine which T-stage is appropriate based on the information given in the radiology report.
Your label options are the following: T1a / T1b / T1c / T2a / T2b / T3 / T4
\\
\noindent\rule{\linewidth}{1pt} 

\section{Single-Prompt Prompt}
\begin{figure*}[t]
\centering
\includegraphics[width=\textwidth]{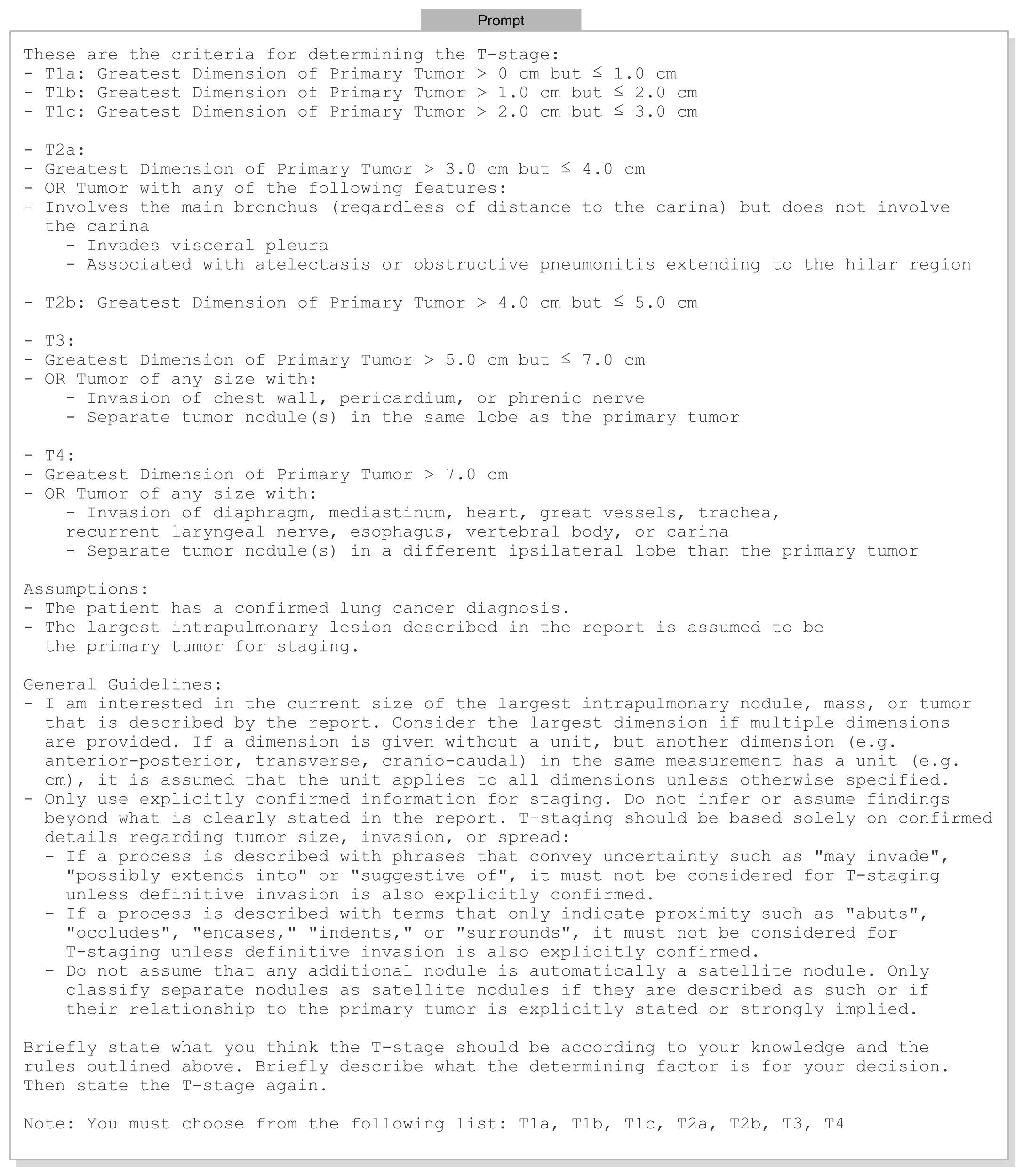}
\caption{Prompt that has been used in the alternative single-prompt approach.}
\label{fig_global_prompt}
\end{figure*}

The prompt is depicted in Figure \ref{fig_global_prompt}.

\bibliographystyle{elsarticle-num}
\bibliography{bib}

\end{document}